\documentclass[conference]{IEEEtran}
\IEEEoverridecommandlockouts
\usepackage{cite}
\usepackage{booktabs}
\usepackage{amsmath,amssymb,amsfonts}
\usepackage{algorithmic}
\usepackage{graphicx}
\usepackage{textcomp}
\usepackage{xcolor}
\usepackage{orcidlink}
\usepackage[moderate,tracking=normal]{savetrees}

\def\BibTeX{{\rm B\kern-.05em{\sc i\kern-.025em b}\kern-.08em
    T\kern-.1667em\lower.7ex\hbox{E}\kern-.125emX}}

\begin{document}

\title{ProCause: Generating Counterfactual Outcomes to Evaluate Prescriptive Process Monitoring Methods\\
}

\author{
Jakob De Moor\textsuperscript{1}\orcidlink{0009-0006-4788-5346}, 
Hans Weytjens\textsuperscript{1,2}\orcidlink{0000-0003-4985-0367}, 
Johannes De Smedt\textsuperscript{1}\orcidlink{0000-0003-0389-0275}
\thanks{This work was supported by the Research Foundation - Flanders (11A6J25N and G039923N), and Internal Funds KU Leuven (C14/23/031).}
\\
\textsuperscript{1}\textit{Research Centre for Information Systems Engineering (LIRIS), KU Leuven, Leuven, Belgium} \\
\textsuperscript{2}\textit{School of Computation, Information and Technology, Technical University of Munich (TUM), Munich, Germany} \\
\{jakob.demoor, hans.weytjens, johannes.desmedt\}@kuleuven.be, hans.weytjens@tum.de
}

\maketitle

\begin{abstract}
Prescriptive Process Monitoring (PresPM) is the subfield of Process Mining that focuses on optimizing processes through real-time interventions based on event log data. Evaluating PresPM methods is challenging due to the lack of ground-truth outcomes for all intervention actions in datasets. A generative deep learning approach from the field of Causal Inference (CI), RealCause, has been commonly used to estimate the outcomes for proposed intervention actions to evaluate a new policy. However, RealCause overlooks the temporal dependencies in process data, and relies on a single CI model architecture, TARNet, limiting its effectiveness. To address both shortcomings, we introduce ProCause, a generative approach that supports both sequential (e.g., LSTMs) and non-sequential models while integrating multiple CI architectures (S-Learner, T-Learner, TARNet, and an ensemble). Our research using a simulator with known ground truths reveals that TARNet is not always the best choice; instead, an ensemble of models offers more consistent reliability, and leveraging LSTMs shows potential for improved evaluations when temporal dependencies are present. We further validate ProCause's practical effectiveness through a real-world data analysis, ensuring a more reliable evaluation of PresPM methods.
\end{abstract}


\section{Introduction}\label{sec:introduction}
Prescriptive Process Monitoring (PresPM) leverages (causal) machine learning (ML) on event logs to establish policies for recommending case-specific interventions during business process executions. These interventions can include managerial escalations, customer interactions, or discounts, all aimed at optimizing Key Performance Indicators (KPIs) such as improving order delivery times or enhancing customer satisfaction scores \cite{bozorgi2023CI}.

Despite their potential, one of the most challenging aspects of developing PresPM methods is evaluating them. Most methods rely on historical event logs, which lack counterfactual outcomes, i.e., the results of intervention actions not taken or recorded. Method performance has to be estimated, since not every prescribed action for a test case has been observed. This issue is known as the \textit{fundamental problem of Causal Inference} (CI) \cite{holland1986}. One common way to approximate real-world performance is by fitting separate models on the historical data to generate counterfactual outcomes. These outcomes serve as a stand-in for ground truth when evaluating a PresPM method. This evaluation approach is extensively used in PresPM, particularly through the RealCause evaluation method \cite{realcause2020}. 
However, RealCause was designed for static intervention optimization and does not account for any temporal dependencies in process data, where the timing, sequence, and evolution of events can significantly influence intervention decisions. Additionally, RealCause relies solely on a single CI learner architecture, TARNet, which limits its effectiveness. It may only provide reliable performance estimates for (parts of) datasets that align with TARNet’s strengths, e.g., when causal effects do not extremely differ between units. This is especially challenging because those strengths depend on the underlying structure of causal effects in a dataset, which is inherently unknowable due to the fundamental problem of CI.

To address these limitations, we introduce Procause, a generative approach designed to estimate performance of PresPM methods in a realistic setting. ProCause enables evaluation using either non-sequential or sequential models, such as LSTMs, as the base model for three key learner architectures from the CI field: S-Learner, T-Learner, and TARNet. Additionally, ProCause includes an ensemble method that averages counterfactual outcomes of these three architectures. This provides a more reliable and less risky default option for stable evaluation, which is particularly useful as the fundamental problem of CI, i.e., the absence of historical counterfactuals, makes it hard to know which learner architecture is more likely to accurately evaluate. We conduct an in-depth assessment of ProCause using a simulator with known ground truths and find that enabling the ensemble option yields the most accurate evaluation approximations on average. Finally, we demonstrate the practical application of ProCause through a real-world data analysis. This involves conducting statistical tests to assess how well the generated factual outcomes align with the observed ones. This comprehensive experiment illustrates the effectiveness of ProCause as a tool for assessing PresPM methods with more confidence given the ground-truth baseline experiments align with real-life performance.

The paper is structured as follows. Section \ref{sec:background} provides background, and Section \ref{sec:motivation} outlines the motivation for ProCause. Section \ref{sec:methodology} describes the methodology, followed by Sections \ref{sec:experimental} and \ref{sec:results}, which cover the experimental setup and results. Section \ref{sec:discussion} offers a discussion, and Section \ref{sec:conclusion} concludes the paper.

\section{Background}\label{sec:background}

\subsection{Prescriptive Process Monitoring \& Causal Inference}
\subsubsection{General}
PresPM is an emerging extension of Process Mining, a field dedicated to discovering, monitoring, and improving real-life processes~\cite{vanderAalst2009}. PresPM advances this goal by recommending case-specific interventions, aiming to directly optimize process KPIs. While still in its infancy, the field is steadily gaining momentum~\cite{branchi2022,bozorgi2023CI,shoush2021,shoush2024}.

One key research direction in PresPM involves leveraging techniques from CI. The objective is to estimate the effect of an intervention using offline observational data. This avoids the need for potentially costly and risky randomized experiments, such as A/B testing a loan assignment strategy in a bank or setting up a randomized controlled trial (RCT). Causal effects of interest may include, for example, the average treatment effect (ATE) at the population level or the conditional average treatment effect (CATE) at the subgroup/individual level, which is the quantity of primary interest in PresPM. Examples of CI-based methodologies in PresPM can be found in \cite{bozorgi2023CI,shoush2021,weytjens2023}. These approaches typically adapt CI approaches to the process context, either by aggregating event data for use with standard CI models \cite{bozorgi2023CI}, or by leveraging sequential models like LSTMs \cite{weytjens2023}. Reinforcement Learning (RL) also offers promising methods for PresPM, but it typically requires online learning, and is thus beyond the scope of this study, which focuses on offline observational data.

While observational data offers clear advantages, it also presents challenges. Since the data reflects an existing policy, e.g., a bank’s current loan strategy, treatment assignment is not random, unlike in controlled trials. A non-random treatment assignment policy can complicate causal estimation. For example, there may be imbalance between treated and untreated groups, or confounding variables that simultaneously influence both treatment and outcome \cite{shalit2017}. Consider for instance a bank offering optional financial literacy workshops (treatment) to improve loan repayments (outcome); individuals who chose to attend might already have had a history of responsible financial behavior, making it difficult to disentangle the workshop's effect from pre-existing tendencies (confounders). Confounding has long been a central concern in studies on CI \cite{shalit2017,curthreally}.

\subsubsection{Causal Inference Learners}
\begin{figure}
    \centering
    \includegraphics[width=0.475\textwidth, height=0.15\textwidth]{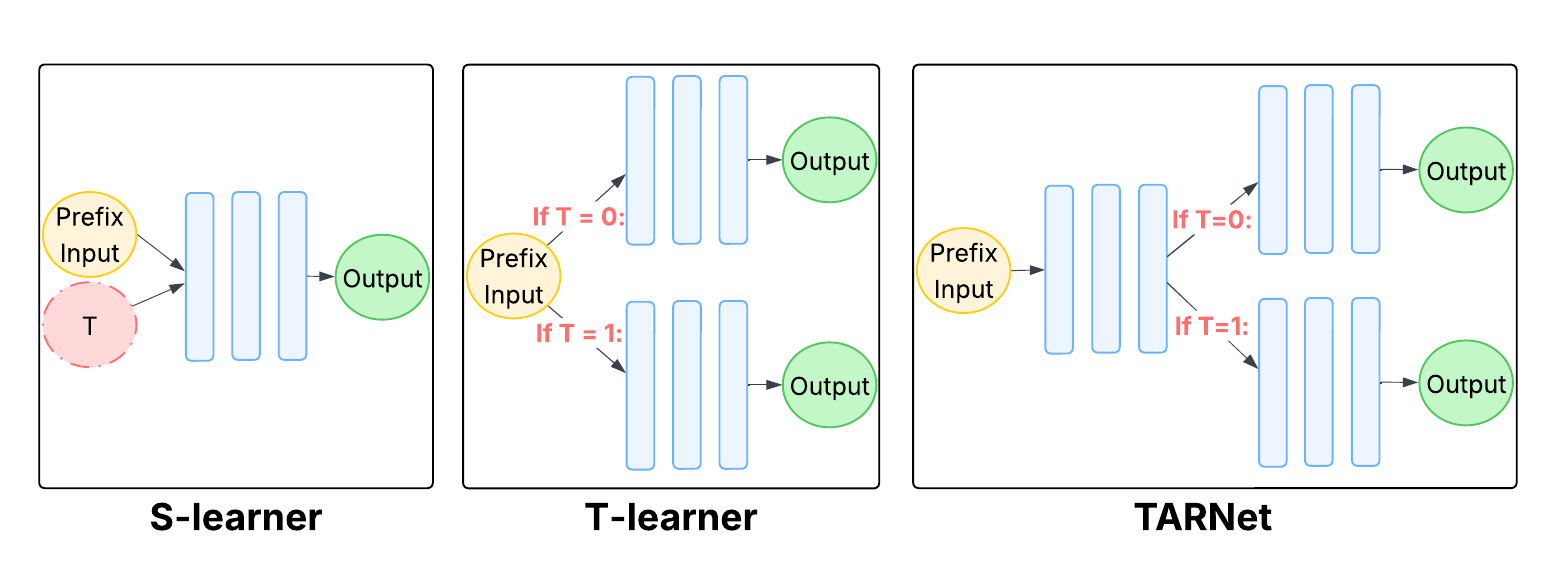}
    \caption{S-learner, T-learner, and TARNet architectures using neural networks; neural network layers are shown in blue.}
    \label{fig:learners}
\vspace{-15pt}
\end{figure}

CI learners are designed to address the challenges of observational data in estimating causal effects. Broadly, they fall into two categories: model-agnostic and model-based. Model-agnostic learners, or meta-learners, define general estimation strategies that are independent of the base model family (e.g., neural networks) they are implemented with. A common example is the S-learner, which fits a single model that predicts the outcome using the treatment as an input in addition to other features. The causal effect is estimated by predicting the outcome under the different possible treatment levels. This method works well when treatment effects are simple or small but can struggle in high-dimensional settings since the treatment signal in the model is weak. The T-learner, on the other hand, trains separate models for each treatment level. It performs well when treatment effects vary significantly across units or when outcome behavior differs sharply between groups. However, it can underperform when data for some treatment levels is sparse \cite{kunzel2017}. Model-based learners require specific model architectures. One example is TARNet, a neural network (NN) architecture that consists of an initial shared representation before branching into separate output layers for each treatment \cite{shalit2017}. This design blends features of NN-based S- and T-learners: like an S-learner, it shares information across treatments (in earlier layers), but like a T-learner, it separates outputs by treatment group. However, this setup may still perform worse \cite{crabbé2022}, for example when treatment groups are highly heterogeneous (favoring T-learners) or severely imbalanced (where S-learners might be more robust). Figure \ref{fig:learners} illustrates the three learners, with NNs used as the base model for both the S-Learner and T-Learner.

\subsection{Evaluating Causal Models}
A central challenge in applying causal analysis to offline observational data lies in the evaluation of the model itself. Consider an observational event log that is split into training and test sets. After training, the PresPM model may estimate that intervening at the second event in a test case would significantly improve the outcome. However, in the observed data, the intervention actually occurred at the third event. Only the outcome of this actual intervention, the factual, is available, while the outcome of the alternative intervention at the second event remains unobservable, known as the counterfactual. This limitation is termed the fundamental problem of CI \cite{holland1986}.

In the CI literature, two main classes of approaches have been proposed to address this problem \cite{realcause2020}. Both aim to provide access to a data-generating process (DGP)—the mechanism that specifies how data is generated in terms of covariates (i.e., observed variables that describe the units or cases), treatments (interventions applied), and outcomes—that enables access to ground-truth counterfactuals for accurate evaluation of causal models.

\subsubsection{Simulated Ground-Truth Generation}
One common approach is to specify the DGP manually, either by generating everything from scratch or by using real covariates and simulating treatments and outcomes. This has the benefit of known DGPs, which makes counterfactuals straightforward to generate and allows for fully accurate evaluations. It also enables complete experimental control by simply adjusting the DGP. While simulated DGPs are widely used in CI \cite{curthreally}, they are rare in PresPM: to our knowledge, only two examples exist \cite{weytjens2023,JDM2025}, both are prior works of the authors, including the simulator used in our experiments below. Some other works, such as \cite{branchi2022}, use forms of simulations as well, but their DGPs rely on existing datasets (and treatment assignment policies), making test counterfactuals possibly inaccurate.
However, simulated DGPs can be unrealistic. Additionally, they may be unintentionally biased in favor of certain algorithms, undermining credibility, especially if used to evaluate a newly proposed method. Strong performance on synthetic data does not necessarily generalize to real-world settings \cite{curthreally}. Designing DGPs is also inherently challenging: they must be simple enough to yield interpretable results, yet complex enough to pose a meaningful problem. This balance becomes even harder in sequential settings, which may explain the scarcity of simulated DGPs in PresPM.

\subsubsection{Model-Driven Ground-Truth Generation}
Another approach is to fit a separate generative model to real data to generate counterfactuals to serve as ground truth. This aims to create more realistic DGPs than purely synthetic setups, while maintaining experimental control by making the generative model adhere to specified requirements. Like CI estimators used in PresPM methods, these generative models often use learner architectures (e.g., TARNet), but instead of estimating intervention effects, they model the distribution of counterfactuals for evaluating CI estimators. To avoid confusion, we refer to such external models used for generating ground truth as \textit{evaluators}. In CI, several examples follow this approach \cite{parikh2022credence,manela2024frugalflows}. An interesting variant is RealCause \cite{realcause2020}, which enables this experimental control due to its easy manipulation of the DGP with `knobs' for confounding, effect heterogeneity, and effect scale. What makes RealCause stand out is the thorough validation of the method in the original paper, which places strong emphasis on the realism of the generated counterfactuals, supported by extensive statistical testing, and its (partial) focus on the CATE, the key quantity of interest in PresPM. Other model-driven evaluation research focuses more on matching user-defined functions than on realism, and on the ATE rather than the CATE. For these reasons, RealCause has been extensively used in PresPM method evaluation \cite{bozorgi2023CI,bozorgi2023RL,shoush2024,shoush2022,shoushwhitebox} where traditional simulation setups are often too complex as mentioned above.
Nevertheless, this approach remains subject to limitations, as the reliability of the evaluation depends on the quality of the fitted model. If the model fails to accurately estimate counterfactuals or algorithm rankings, the resulting evaluations may be misleading \cite{curthreally}.

\section{Motivation}\label{sec:motivation}

\subsection{Current Limitations}
Despite the mentioned benefits of RealCause, the method has two main limitations. First, RealCause relies solely on multi-layer perceptrons (MLP), which require encoding schemes for process data, aggregating the sequential information that exists over time (e.g., aggregation encoding). This can lead to information loss, particularly in the presence of long-range dependencies or when event order is crucial \cite{dlppm2023}. 

Second, it only implements the TARNet architecture, which may not yield reliable counterfactual estimates across all datasets, as described in Section \ref{sec:background}. More critically, selecting the most appropriate learner depends on identifying certain data characteristics (e.g., effect heterogeneity, see Section \ref{sec:background}), which is inherently difficult and would require strong assumptions of the underlying DGP as the individual treatment effects cannot be observed due to the fundamental problem of CI. For instance, it is challenging to observe whether the treatment effect is simple enough for an S-learner or varies significantly across units, which would favor a T-learner. This challenge is amplified in the context of PresPM. Existing causal learners have primarily been studied in static settings, whereas process data is dynamic and temporally complex. Even if one can identify suitable learners in static scenarios, this becomes more difficult in sequential contexts. For example, the timing of a treatment during a business process adds another layer of complexity for the decision of the learner: strong causal effect differences may occur not just between individual cases (which would favor the T-learner), but also within cases due to effect differences because of treatment timing. 

Because of the fundamental problem of CI, evaluation in PresPM is especially critical. Its reliability and performance directly shape how PresPM methods are developed, compared, and trusted. Therefore, the limitations discussed highlight an opportunity for a more robust evaluation approach.

\subsection{Contributions}
Motivated by these concerns, we propose ProCause, a method for process-aware, realistic and reliable validation of PresPM methods. Our contributions are as follows:

\begin{itemize}
     \item[a)] We extend current CI architectures to support both non-sequential and sequential models, particularly LSTMs.\footnote{We use LSTMs as a representative sequential model due to their proven effectiveness in process tasks \cite{dlppm2023}. While alternatives like transformers exist, our focus is on demonstrating how any sequential model can be integrated into the CI evaluation framework.}
    \item[b)] We extend the evaluation framework introduced by RealCause using TARNet by incorporating additional CI learners, specifically S- and T-learners. Crucially, we also combine these into an ensemble and show that this approach yields the most reliable evaluation of PresPM methods when little is known about the underlying DGP. Since selecting the best learner typically requires assumptions of the DGP due to the fundamental problem of CI, we demonstrate that the ensemble offers a robust, low-risk default, as different learners can better estimate different parts of the distribution of interest.
\end{itemize}

\section{Methodology}\label{sec:methodology}

\subsection{Preliminaries}
\subsubsection{Event Log Data}
PresPM research focuses on event log data, which captures records of process executions. Here, a case refers to a single instance of a process execution and consists of a series of events. Each event represents the execution of an activity and includes three key attributes: a case identifier, the activity label, and a timestamp. Events may also have additional attributes, such as the resource performing the activity. A trace is the sequence of all events within a case. A prefix is any initial segment of a trace leading up to a particular event in the trace, representing a partial process execution.

\subsubsection{Conditional Average Treatment Effect}
Let \( T \) be a categorical scalar random variable denoting the treatment, taking values of the set \( \{0, 1, \ldots, K\} \), where \( K \geq 1 \). Let \( X \) be a set of random variables corresponding to the observed covariates of a prefix, and let \( Y \) be a scalar random variable denoting the outcome of interest. Furthermore, let \( Y(t) \) denote the potential outcome that would be observed if the treatment were set to level \( t \). We consider the evaluation of PresPM methods that estimate the CATE $\tau_{t,t'}(x)$ between any two treatment levels of interest \( t \) and \( t' \) on a given prefix $x$, defined as follows:

\vspace{-5pt}
\begin{equation}
    \tau_{t,t'}(x) = \mathbb{E}[Y(t) - Y(t') \mid X = x]
\end{equation}
\vspace{-5pt}

The CATE can be identified from observational data under a set of strong and inherently untestable assumptions: \textit{ignorability}, \textit{positivity}, \textit{consistency}, and \textit{no interference}. For detailed discussions of these assumptions, we refer the reader to \cite{imbensrubinsCI}.

\subsection{ProCause}
We begin by outlining the general setup, which follows the approach in RealCause at a high-level. We then delve into our implementations, where the key improvements within ProCause emerge.

\subsubsection{General}
Generative evaluation approaches like Realcause aim to approximate the true DGP, represented by the joint distribution \( P(X, T, Y) \). This joint distribution can be factorized via ancestral sampling as \( P(X) \rightarrow P(T \mid X) \rightarrow P(Y \mid T, X) \). For evaluation purposes, we can sample \( P(X) \) directly from real (test) data (as in a semi-synthetic setup), then we can fit generative models for \( P(T \mid X) \), which allows control over the treatment assignment---for example, to vary the level of confounding. Similarly, we can fit a model to the outcome distribution \( P(Y \mid T, X) \), enabling the generation of counterfactual outcomes (given a treatment and prefix) to evaluate PresPM methods. When needed, we can sample from the full joint distribution using this ancestral structure.

The generative models take covariates \( X \) as input and output the parameters for a predefined probability distribution type. For each input prefix \( X \), the models may produce different parameters, though the type of distribution remains fixed. For binary data, this is typically a Bernoulli distribution, where the model predicts the mean. For real-valued data, a Gaussian distribution is common, with the model outputting both the mean and log-variance. For more complex cases, flexible distributions like sigmoidal flows can be used \cite{huang2018neural}. When modeling mixed variables—those with both discrete and continuous components—a weighted combination of a continuous distribution and predefined discrete points is used. While specifying the distribution type in advance may seem limiting, the flexibility of sigmoidal flows offers a practical solution when no clear prior knowledge is available.

\subsubsection{Implementations}
This is where our approach diverges significantly from RealCause. Instead of solely relying on TARNet as done by RealCause, we also use S-learners and T-learners to model the outcome distributions. Additionally, we offer an \textit{ensemble} option that averages samples from the outcome distributions estimated by all three learners. Each of these learners is built using NNs, as displayed in Figure \ref{fig:learners}. Unlike Realcause, which offers only one option, users can choose between two types: MLPs or LSTMs. For the S-learners and T-learners, the entire network can be composed of either MLPs or LSTMs. In the case of TARNet, if MLPs are selected, the architecture follows the RealCause setup, where all layers are MLPs. If LSTMs are chosen, we modify TARNet by incorporating LSTMs in the shared representation layers. This allows the model to capture temporal dependencies in the sequential input data. We then use MLPs in the subsequent treatment-specific layers, as the temporal structure has already been encoded by the shared LSTM. Using LSTMs only in the shared representation layer reduces model complexity and helps prevent overfitting in the treatment-specific layers, especially given limited data per treatment group. In low-data settings, LSTMs can be harder to train and more prone to overfitting, making simpler architectures more suitable for these treatment-specific components. For estimating the treatment assignment distribution $P(T \mid X)$, we use a separate standard MLP or LSTM, independent of the choice of CI learner.

\section{Experimental Setup}\label{sec:experimental}
In this section we describe our experimental setup to assess ProCause’s ability to evaluate PresPM methods. \footnote{The complete code for these experiments is openly accessible at \url{https://github.com/JakobDeMoorKULstudent/ProCause}.} We first discuss the preprocessing used for all setups. Then we delve deeper into our simulation setup, which is absolutely necessary to completely accurately assess performance due to the fundamental problem of CI. Simulations also allow us to vary key parameters in the CI setting, such as the level of confounding. However, since simulations may not reflect real-world complexity, we complement this with statistical tests on two real-world datasets to assess the realism of the estimated distributions. Our experiments primarily focus on ProCause’s ability to estimate the outcome distribution. \footnote{The treatment distribution is approximated in the same way as in RealCause, including the flexible tuning `knobs' (see \cite{realcause2020}).} Like before, we refer to models used to evaluate PresPM methods as evaluators. Note that while evaluators for PresPM have been validated before \cite{bozorgi2023CI, JDM2025}, this pipeline delivers the most comprehensive assessment in PresPM thus far.

\subsection{Preprocessing}
Cases are truncated at predefined intervention points, assumed known as in standard PresPM practice, yielding prefixes that end just before intervention or, if untreated, at the latest point to ensure the outcome is also learned under no treatment. For MLPs, we use last-event encoding for time features and aggregation encoding for others. LSTMs use tensor encoding \cite{ppmremainingtime2019}, processing full event sequences with case-level features added at the final layer. Categorical variables are one-hot encoded (6 most frequent levels and “other”, except for the activity label, which is uncapped). Continuous features are standardized.

\subsection{Simulator: Accuracy of the Evaluation}
We use SimBank as our simulator \cite{JDM2025}. While using multiple simulators would offer more robust results, to the best of our knowledge, SimBank is the only simulator specifically designed for PresPM. It models a loan application process in a bank and has been thoroughly validated. The KPI to optimize is the total profit made on loans in a test set. We focus on the two most complex interventions in SimBank: setting the correct interest rate (\textit{set interest rate}), having the option between 3 levels implemented at a fixed intervention point, and choosing the right moment to contact HQ (\textit{time contact HQ}), which adds a crucial timing element. Details of these interventions are provided in \cite{JDM2025}. SimBank also allows us to vary confounding levels by adjusting the percentage of confounded data compared to RCT (i.e., unconfounded) data in the training set. As previously shown \cite{bozorgi2023CI,JDM2025}, increased confounding typically degrades a PresPM method’s performance on average. We vary confounding here as well across 6 levels, ranging from 75\% to 99.9\% of the training data being confounded, to (1) study how it affects an evaluator’s ability to assess PresPM methods, and (2) ensure evaluators are tested on a wider range of DGPs, confirming that any demonstrated strengths by the evaluators hold consistently across different scenarios. These levels mirror real-world observational data (typically heavily confounded), while still varying DGPs to thoroughly assess ProCause.

\begin{figure}
    \centering
    \includegraphics[scale=0.235]{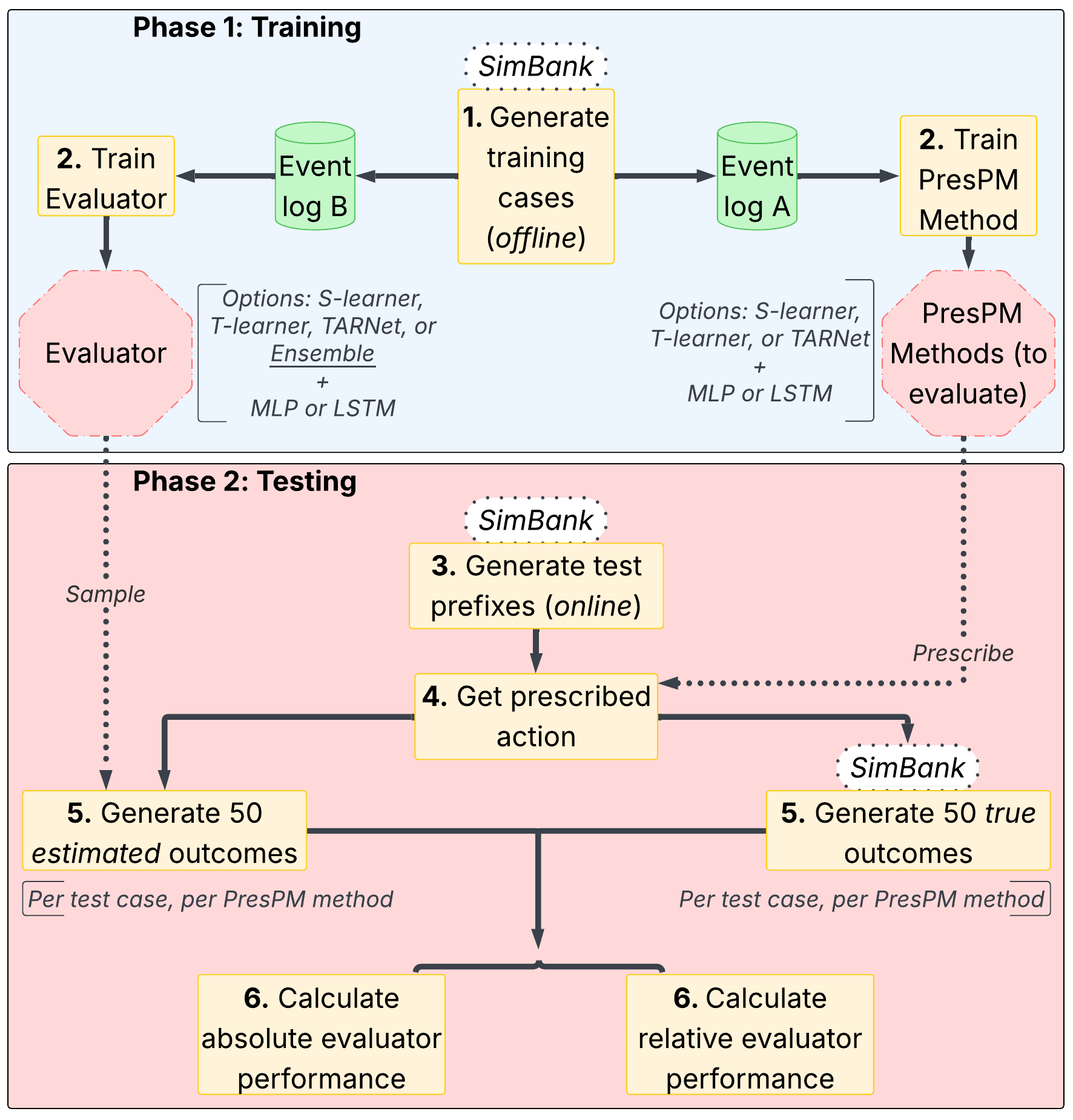}
    \caption{The experimental setup of our assessment of ProCause using a simulator. This pipeline is repeated for every intervention and confounding level.}
    \label{fig:setup}
\vspace{-10pt}
\end{figure}

The simulator-based assessment compares ground-truth evaluations of PresPM methods with those from evaluators. We design a pipeline that reflects realistic ProCause usage for evaluating and comparing PresPM methods. This setup is repeated across intervention types and confounding levels, and consists of two main stages: training and testing. A summary can be seen in Figure \ref{fig:setup}.

\subsubsection{Training}
Here we outline the training of both the evaluators and the PresPM methods to evaluate with those evaluators. 

\paragraph{Evaluators}
We begin by generating a fixed training set using SimBank, which we split into two parts. One subset of 10,000 cases is used to train evaluators. Each evaluator is defined by a choice of learner (including an ensemble, which involves training 3 separate learners) and an underlying model (either an MLP or an LSTM). Each configuration is trained with 5 different random seeds, and all subsequent results are averaged over these runs. As mentioned in section \ref{sec:methodology}, the treatment and outcome distribution types must be specified beforehand. For SimBank, we assume a mixed distribution (discrete points and sigmoid flow) for the outcome, a Bernoulli distribution for the treatment in the \textit{time contact HQ} intervention, and a categorical treatment distribution for \textit{set interest rate}. The models are then trained by maximizing the log-likelihood using the Adam optimizer. The training set is temporally split, and we use 20\% as a validation set for hyperparameter tuning (hidden layer dimension, learning rate, and batch size) and for early stopping to prevent overfitting. For every setup—defined by the combination of learner, underlying model, confounding level, and intervention—we apply the same hyperparameter tuning procedure using Bayesian optimization. The maximum possible network size resulting from this tuning is identical across setups. This ensures a fair comparison, preventing any implementation from benefiting from a larger model.

\paragraph{PresPM Methods to Evaluate}
The remaining 2,500 cases are used to train the PresPM methods we wish to evaluate with ProCause, since the focus is on ProCause’s ability to assess given PresPM methods. These include a random policy and six informed PresPM methods, each based on different learner-model combinations (excluding ensembles). Training and preprocessing follow the same procedures as those used for the evaluators, but with Mean Squared Error (MSE) on the profit as loss function and using 5 random seeds. A PresPM method is then defined by one estimator (learner-model combination) and provides prescriptions as follows: given a prefix, the estimator predicts the outcomes under all possible treatment levels and selects the action corresponding to the highest expected outcome. For the \textit{time contact HQ} intervention, which requires precise timing, we tune a threshold on a separate validation set (20\%). The intervention is then triggered if the estimated effect of intervening now versus waiting exceeds this threshold. 

\subsubsection{Testing}
The second component concerns the actual assessment of ProCause as an evaluator of PresPM methods.

\paragraph{Outcome Generation}
Once we have both the evaluator(s) and the PresPM methods to assess, we want to test whether the evaluator can accurately estimate the true outcome distribution \( P(Y \mid T, X) \) given a PresPM method policy (which provides \( T \)) and test prefix \( X \), across a full test set. 
To do this, we first initiate 1,000 test cases using SimBank. At each possible intervention point in a test case, we check the actions prescribed by the PresPM methods. Once a PresPM method decides on an intervention—or if no intervention is made and no further intervention points remain—we obtain samples from the true local outcome distribution \( P(Y \mid T, X) \) by generating the rest of the case using 50 different random seeds with SimBank and saving the outcomes. Samples from the estimated outcome distribution are obtained similarly: given a PresPM method intervenes (or never intervenes), we generate 50 estimated outcomes using 50 different seeds when sampling from the distribution defined by the outputted parameters of the evaluator. This results in 1,000 × 50 samples per PresPM method for both the true and estimated outcome distributions, giving a reliable approximation of the local outcome distribution \( P(Y \mid T, X) \). 
We use 50 samples per case to reliably approximate the local outcome distributions, while keeping the overall computational load manageable. \footnote{In total, this amounts to 2 (each intervention) $\times$ 6 (confounding levels) $\times$ 7 (PresPM methods) $\times$ 1,000 (cases) $\times$ 50 samples, not considering evaluator and PresPM method iterations.}

\paragraph{Metrics}
Now that we have collected, for each test case and PresPM method combination, 50 samples from both the true and estimated outcome distribution, we can assess whether a practitioner using ProCause to evaluate PresPM methods would obtain results that reflect their true performance. We define a good evaluator by two main qualities. The first is the ability to accurately estimate the outcome distribution for a PresPM method applied to a test case, which we refer to as absolute performance, measuring an evaluator's ability in evaluating the impact of a \textit{single} PresPM method. The second is the ability to correctly rank \textit{multiple} PresPM methods across a test set, which we call relative performance.

\textit{Absolute evaluator performance} focuses on how well the estimated outcome distribution \( P(Y \mid T, X) \) matches the true distribution for each test case, given a PresPM method which provides \( T\). We quantify this using the Wasserstein-1 distance (WD) between the 50 outcomes drawn from the true distribution and the 50 from the estimated one. The metric is then averaged over all test cases. The Wasserstein-1 distance in 1 dimension between two distributions represented by 50 samples each is the average absolute difference between their sorted values ($x_{(i)}$ and $y_{(i)}$): $W_1(\mu, \nu) = \frac{1}{50} \sum_{i=1}^{50} \left| x_{(i)} - y_{(i)} \right|$.
To reflect practical relevance, we average the WD across the informed (non-random) PresPM methods, i.e., the ones a practitioner would realistically consider. We also report the distance separately for the random policy. While less relevant in practice, the random policy introduces the highest diversity in prescribed interventions. This makes it useful for testing the evaluator’s robustness across a wider space of possible inputs. Informed PresPM methods, on the other hand, are likely to make consistent treatment choices for similar prefixes, and thus exercise in a narrower part of the evaluator’s input space.

\textit{Relative evaluator performance} refers to the evaluator’s ability to correctly rank a set of PresPM methods across a test set. For each test case and PresPM method, we compute the average outcome over the 50 samples, then sum these averages across all test cases to obtain the total profit of each PresPM method (and random policy). This process is performed using both the true and estimated outcomes, resulting in two rankings: one based on the true profit and one based on the evaluator’s estimates.
To assess how well the evaluator captures the correct ranking, we calculate the Kendall's rank correlation coefficient (Kendall's \(\tau\)) between the true and estimated PresPM method rankings. This correlation measures the agreement between the two orderings, with higher values indicating better evaluator performance. Kendall's \(\tau\) is defined as $\tau = \frac{C - D}{\frac{1}{2} n (n-1)}$ where \(C\) is the number of concordant pairs, \(D\) is the number of discordant pairs, and \(n\) is the total number of observations.

\subsection{Real datasets: Realism of the Evaluation}
In this component, we assess whether ProCause’s generated outcomes and treatment assignments align with real-world data. We use the BPIC12 and BPIC17 datasets \cite{bpic12,bpic17}, where the intervention involves a timed intervention of sending a second application offer \cite{bozorgi2023CI} and the outcome is binary, with 1 meaning the application is approved/pending. We use the same preprocessing and training setups of the simulator experiment, using a temporal split to setup train, validation, and test sets, and assuming Bernoulli distributions for both outcome and treatment. To evaluate realism, we use a setup that is consistent with those used in RealCause \cite{realcause2020} and PresPM literature \cite{bozorgi2023CI}. We generate outcome and treatment variables on the test set and compare it with the real test set using statistical tests that check whether two datasets (generated and real) come from different distributions. For univariate analysis, we use the Kolmogorov–Smirnov (KS) test and the Epps–Singleton (ES) test where the latter is more appropriate for discrete variables like binary treatments or outcomes. For multivariate analysis, we use the Friedman–Rafsky test, k-nearest neighbor (kNN) test, and the energy test. Additionally, we run permutation tests with Wasserstein-1 and Wasserstein-2 distance metrics.
\footnote{In RealCause, a multivariate test is also applied to \( T \), \( Y \), and attributes \( X \). Since ProCause is used as an evaluator, end users typically apply it with the real \( X \), making this test less relevant in our context. For completeness, we ran it using generated \( T \) and \( Y \) with real \( X \). But due to \( X \)'s high dimensionality, the test loses power and yields p-values near 1, making the generated data appear highly realistic, though this is expected and not especially informative.} 
The null hypothesis is that the generated and original data share the same distribution. High p-values (e.g., $p > 0.1$) suggest no significant difference between them.

\section{Results}\label{sec:results}
This section presents the results. We start with the simulator to assess evaluation accuracy, discussing the impacts of learner type, model type, and providing general remarks. We conclude with real datasets to validate the realism of the generated variables.

\begin{figure*}[t]
    \centering
    \includegraphics[width=\textwidth]{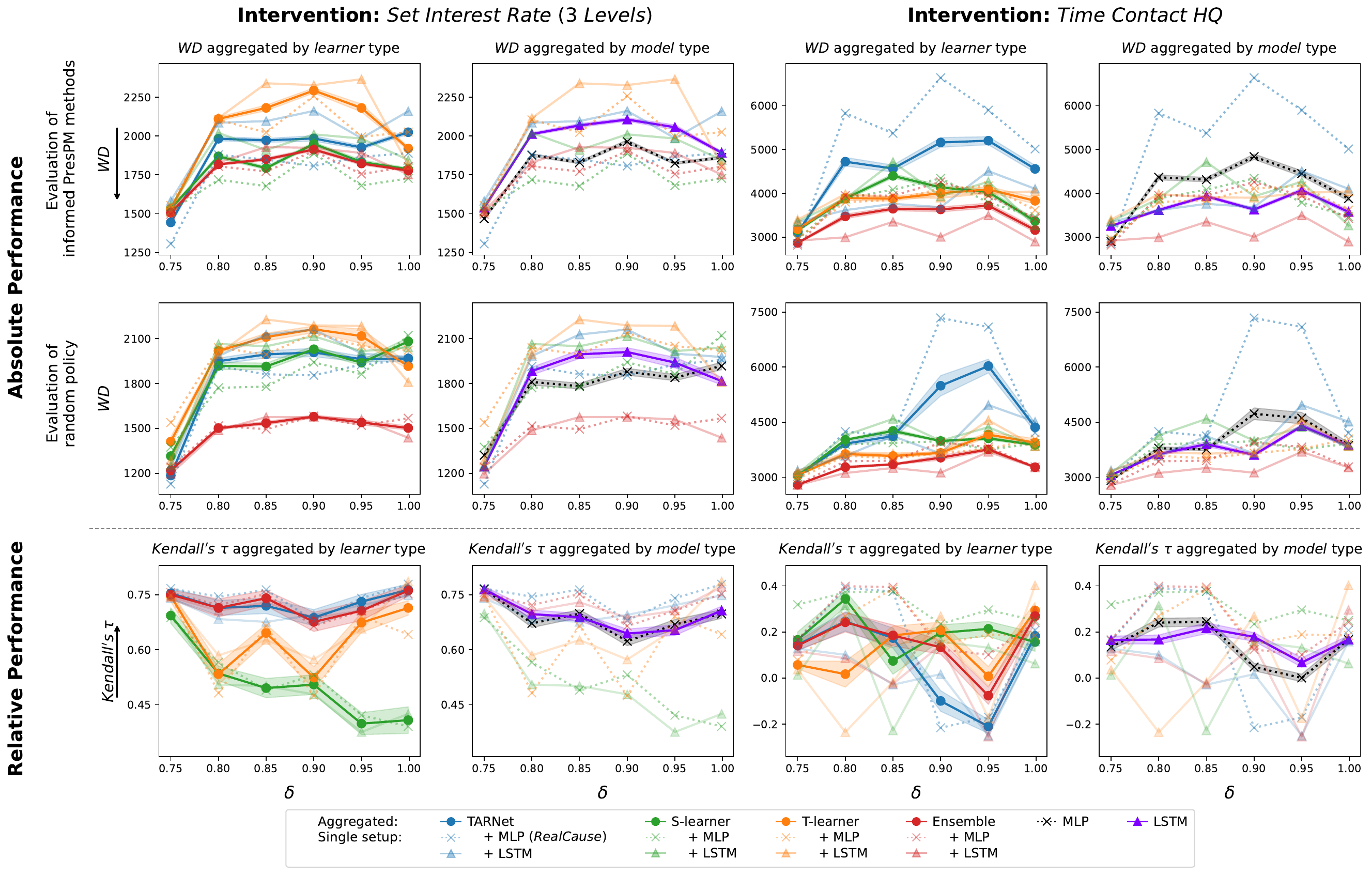}
    \caption{The results of our assessment of ProCause using a simulator, displaying both absolute performance in WD (lower is better), and relative performance using Kendall’s Tau (higher is better). We aggregate per learner and model type to identify the sources of gain, where RealCause corresponds to TARNet + MLP. $\delta$ is the level of confounding. The shaded areas indicate one standard error.}
    \label{fig:results}
\vspace{-10pt}
\end{figure*}

\subsection{Simulator: Accuracy of the Evaluation}
Figure \ref{fig:results} shows the results for the simulator experiment, focusing on the evaluation accuracy of different ProCause configurations. In this figure, we aggregate results by learner and model type to pinpoint the source of performance improvements. Since our contributions involve both the learner and base model aspects of an evaluator, this aggregation helps isolate their effects. We show the non-aggregated results (single setups) here in the background of each plot for clarity\footnote{Full results—including all individual PresPM method rankings—are available at: \url{https://github.com/JakobDeMoorKULstudent/ProCause}.}.
Additionally, we separately display evaluation of non-random (informed) PresPM methods and the random policy: evaluating informed ones reflects the intended use case of the evaluator, while a random policy tests an evaluator's robustness by covering a wider input space (see Section \ref{sec:experimental}). Note that our results focus on (absolute and relative) evaluator performance, not on the PresPM methods. References to a learner or model type specifically denote an evaluator configuration.

\subsubsection{Ensemble versus individual learners}
ProCause introduces a flexible learner configuration that includes S-learners, T-learners, and TARNet, and combines them into an ensemble. Results show that the ensemble learner offers the most consistently strong performance across interventions, confounding levels, and assessment types (absolute and relative). It is especially stable in evaluating random policies, where treatment assignment varies stochastically. Only when ranking PresPM methods for the \textit{time contact HQ} intervention is considered, the ensemble is not the top
performer, though it still remains in the top 2/3. In contrast, individual learners perform well only under specific conditions. For example, RealCause, based solely on TARNet, may provide less reliable evaluations: TARNet ranks PresPM methods well under the \textit{set interest rate} intervention but performs generally poor for the \textit{time contact HQ} intervention. This supports the central motivation for the ensemble: when the DGP is unknown, individual learners can exhibit unpredictable behavior, whereas the ensemble buffers against such variance.

\subsubsection{LSTM versus MLP}
ProCause supports both MLP- and LSTM-based architectures. Within the fixed intervention scenario (\textit{set interest rate}), MLP-based models tend to exhibit slightly more consistent performance compared to LSTM-based models. This suggests that, under conditions with fixed treatment timing, MLPs may offer practical advantages due to their simplicity and stability. For the intervention with time-varying treatment points, such as \textit{time contact HQ}, LSTM models perform slightly better on average. While the improvement is not large, it indicates that temporal modeling has potential value in such settings. As RealCause relies exclusively on MLP-based models, this flexibility in ProCause to include LSTMs allows it to better adapt to a wider range of intervention types, particularly when timing is a relevant factor.

\subsubsection{Additional general observations}
First, when comparing multiple evaluators in the results, it is important to note that absolute performance does not always match relative performance. An evaluator might accurately estimate local outcome distributions but still rank PresPM methods poorly, or vice versa, in comparison with other evaluators. For example, in the \textit{set interest rate} intervention, TARNet shows poor absolute performance but good relative performance, while the S-learner demonstrates the opposite.

A second, related point concerns the impact of confounding on an evaluator’s performance. As confounding increases, absolute evaluator performance tends to decline on average, which mirrors the decline generally seen in PresPM methods themselves \cite{bozorgi2023CI,JDM2025}. However, confounding does not have a consistent effect on relative performance: the quality of rankings does not show a general trend and can be just as good (or bad) with the heaviest confounded data as with the least confounded data considered.

Finally, model rankings under the \textit{set interest rate} intervention are generally in agreement with the true ranking. However, under the \textit{time contact HQ} intervention, there is still room for improvement, as indicated by the Kendall's \(\tau\) showing only weak agreement between the true and estimated rankings.

\subsection{Real datasets: Realism of the Evaluation}
Table \ref{tab:stat_tests} presents the results of univariate and multivariate tests comparing the generated outcome $Y$ and treatment $T$ variables to the real ones of the BPI event logs. The tests were conducted using two ensemble configurations, one with an MLP and the other with an LSTM as the base model. In every case, at least one of these configurations produced a $p$-value above commonly accepted significance thresholds (e.g., $p > 0.1$). This indicates that there is no statistical evidence to suggest a difference between the distributions of the generated and real variables. Consequently, these results support the conclusion that the data generated by ProCause is realistic.

\begin{table}
    \centering
    \fontsize{6}{7}\selectfont
    \caption{Statistical tests to validate realism of generated treatment and outcome variables (ensemble setup).}
    \label{tab:stat_tests}
    \setlength{\tabcolsep}{5pt}
    \begin{tabular}{lcccc}
        \toprule
         & \multicolumn{2}{c}{BPI12} & \multicolumn{2}{c}{BPI17}\\
         \cmidrule(r){2-3} \cmidrule(l){4-5}
             & \multicolumn{1}{c}{MLP} & \multicolumn{1}{c}{LSTM} & \multicolumn{1}{c}{MLP} & \multicolumn{1}{c}{LSTM}\\
        \midrule
        T KS & 0.56 & 0.59 & 0.58 & 0.66 \\
        T ES & 0.66 & 0.45 & 0.00 & 0.56 \\
        Y KS & 0.51 & 0.03 & 0.55 & 0.01 \\
        Y ES & 0.21 & 0.01 & 0.26 & 0.10 \\
        (T,Y) Wass1 & 0.34 & 0.20 & 0.73 & 0.78 \\
        (T,Y) Wass2 & 0.34 & 0.20 & 0.68 & 0.73 \\
        (T,Y) FR & 0.33 & 0.34 & 0.77 & 0.76 \\
        (T,Y) kNN & 0.09 & 0.30 & 0.66 & 0.78 \\
        (T,Y) Energy & 0.45 & 0.27 & 0.75 & 0.81 \\
        \bottomrule
    \end{tabular}
\vspace{-15pt}
\end{table}

\section{Discussion}\label{sec:discussion}
\begin{figure}
    \centering
    \includegraphics[width=0.4\textwidth]{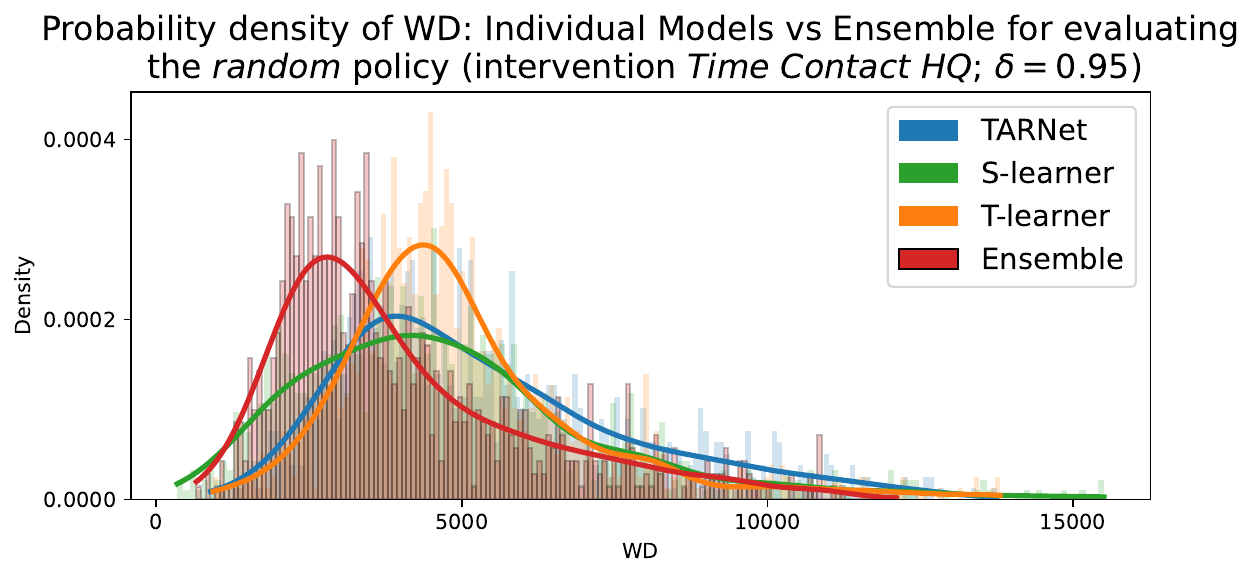}
    \vspace{-7pt}
    \caption{The probability density of the WD over the test set between true and generated outcome distributions by each learner, where bars show the empirical distribution and lines are a smoothed representation, evaluating the random policy for intervention \textit{time contact HQ} at $\delta = 0.95$.}
    \label{fig:ensemble}
\vspace{-15pt}
\end{figure}

Our results show that ProCause, especially its ensemble option, offers a robust and reliable framework for evaluating PresPM methods. By combining multiple learners, the ensemble consistently delivers stable performance across diverse DGPs, reducing sensitivity to specific data traits and balancing individual model weaknesses. As visualized in Figure \ref{fig:ensemble}, which shows the probability density of WDs between true and estimated distributions for the \textit{time contact HQ} intervention at $\delta$ 0.95, the ensemble in ProCause achieves both lower WD values and reduced variability, underscoring its robustness. In contrast, our results indicate that relying solely on TARNet, as in the RealCause setup, does not provide the same level of reliability. Because the true DGP is unknown in practice, ProCause’s ensemble option offers a dependable default for evaluation.

ProCause also extends RealCause by integrating LSTM-based learners. Our results indicate that LSTMs can improve evaluation accuracy for PresPM methods with time-dependent treatments, such as the \textit{time contact HQ} intervention. However, these improvements are not guaranteed across all scenarios. For the fixed intervention (\textit{set interest rate}), MLP-based models tend to be more stable. Therefore, the choice between LSTM and MLP should be guided by the characteristics of the intervention and the expected temporal structure of the data, for example, by first analyzing the predictive performance of both sequential and non-sequential models to identify which one may be advantageous.

As shown in the results, an evaluator’s accuracy in estimating outcomes under a given PresPM policy (its absolute performance) does not always reflect how well it can rank different PresPM methods compared to other evaluators. An evaluator might have poor absolute performance but still rank PresPM methods accurately. This can happen if the absolute errors do not significantly affect the overall ranking, if the errors are consistent across all PresPM methods being evaluated, or if they cancel each other out when aggregated to obtain a PresPM method’s estimated performance. On the other hand, even small but uneven absolute errors across the PresPM methods evaluated can distort the rankings, resulting in strong absolute but poor relative performance. This distinction between absolute and relative performance also explains why, when examining one evaluator’s performance across different levels of confounding, the confounding can impact absolute performance but not necessarily the evaluator’s ability to rank PresPM methods. In some cases, even under high confounding, absolute errors become irrelevant when the goal is to compare PresPM methods.

While our validation showcases ProCause’s promising potential, our results indicate that model-driven evaluation has room for improvement as shown by the weak ranking agreement for the \textit{time contact HQ} intervention. Nevertheless, our results indicate that ProCause provides a more robust and flexible alternative to the widely used RealCause. We especially recommend an ensemble over single models like TARNet, and LSTM-based models when temporal dependencies are suspected to play a significant role.

\section{Conclusion and Future Work}\label{sec:conclusion}
In summary, ProCause advances the evaluation of PresPM methods by integrating both sequential and non-sequential models, along with a broader set of CI learner architectures. Our findings, supported by simulations with known ground truths and real-world data, show that using an ensemble of CI learners improves reliability and using sequential models shows potential to better capture temporal dependencies in process data.
A first key limitation is reliance on a single simulator, though we mitigated this by varying DGPs using different interventions and confounding levels. Secondly, evaluators like ProCause still face the fundamental problem of CI: their real-world performance is unknowable. We tried to address this using a simulator for ground truth testing and statistical tests to assess realism. Lastly, ProCause relies on certain CI assumptions (see Section \ref{sec:methodology}) that, while standard in CI and PresPM, remain inherently untestable.

Future work will focus on improving model-driven evaluations—for instance, exploring whether learners like the X-learner, which output CATE rather than outcomes, can still be leveraged; extending ProCause to support prescriptions for combinations of interventions; and applying it to resource allocation tasks.
Nevertheless, our results show that ProCause offers a promising step towards more comprehensive evaluations of PresPM methods.

\bibliographystyle{splncs04}
\bibliography{references}

\begin{thebibliography}{10}
\providecommand{\url}[1]{\texttt{#1}}
\providecommand{\urlprefix}{URL }
\providecommand{\doi}[1]{https://doi.org/#1}

\bibitem{vanderAalst2009}
van~der Aalst, W.M.P.: Process Mining, pp. 2171--2173. Springer US (2009). \doi{10.1007/978-3-662-49851-4}

\bibitem{bozorgi2023RL}
Bozorgi, Z.D., Dumas, M., Rosa, M.L., Polyvyanyy, A., Shoush, M., Teinemaa, I.: Learning when to treat business processes: Prescriptive process monitoring with causal inference and reinforcement learning. In: Advanced Information Systems Engineering. vol. 13901. Springer (2023)

\bibitem{branchi2022}
Branchi, S., Di~Francescomarino, C., Ghidini, C., Massimo, D., Ricci, F., Ronzani, M.: Learning to act: A reinforcement learning approach to recommend the best next activities. In: BPM Forum. vol.~458. Springer (2022)

\bibitem{crabbé2022}
Crabb{\'e}, J., Curth, A., Bica, I., van~der Schaar, M.: Benchmarking heterogeneous treatment effect models through the lens of interpretability. In: 36th Conference on NeurIPS Datasets and Benchmarks Track. vol.~35 (2022)

\bibitem{curthreally}
Curth, A., Svensson, D., Weatherall, J., van~der Schaar, M.: Really doing great at estimating cate? a critical look at ml benchmarking practices in treatment effect estimation. In: NeurIPS Track on Datasets and Benchmarks. vol.~1 (2021)

\bibitem{bozorgi2023CI}
{Dasht Bozorgi}, Z., Teinemaa, I., Dumas, M., {La Rosa}, M., Polyvyanyy, A.: Prescriptive process monitoring based on causal effect estimation. Information Systems  \textbf{116} (2023)

\bibitem{JDM2025}
De~Moor, J., Weytjens, H., De~Smedt, J., De~Weerdt, J.: Simbank: from simulation to solution in prescriptive process monitoring. In: BPM Forum (2025), \url{https://arxiv.org/abs/2506.14772}, accepted for publication, to appear

\bibitem{bpic12}
van Dongen, B.: Bpi challenge 2012 (2012). \doi{10.4121/uuid:3926db30-f712-4394-aebc-75976070e91f}

\bibitem{bpic17}
van Dongen, B.: Bpi challenge 2017 (2017). \doi{10.4121/uuid:5f3067df-f10b-45da-b98b-86ae4c7a310b}

\bibitem{holland1986}
Holland, P.W.: Statistics and causal inference. Journal of the American Statistical Association  \textbf{81} (1986)

\bibitem{huang2018neural}
Huang, C.W., Krueger, D., Lacoste, A., Courville, A.: Neural autoregressive flows. In: ICML. vol.~80. PMLR (2018)

\bibitem{imbensrubinsCI}
Imbens, G.W., Rubin, D.B.: Causal Inference for Statistics, Social, and Biomedical Sciences: An Introduction. Cambridge University Press (2015)

\bibitem{kunzel2017}
Künzel, S., Sekhon, J., Bickel, P., Yu, B.: Meta-learners for estimating heterogeneous treatment effects using machine learning. National Academy of Sciences  \textbf{116} (2017)

\bibitem{realcause2020}
Neal, B., Huang, C.W., Raghupathi, S.: Realcause: Realistic causal inference benchmarking (2021), \url{https://openreview.net/forum?id=m28E5RN64hi}

\bibitem{parikh2022credence}
Parikh, H., Varjao, C., Xu, L., Tchetgen, E.T.: Validating causal inference methods. In: 39th ICML. vol.~162. PMLR (2022)

\bibitem{dlppm2023}
Rama-Maneiro, E., Vidal, J.C., Lama, M.: Deep learning for predictive business process monitoring: Review and benchmark. IEEE Transactions on Services Computing  \textbf{16} (2023)

\bibitem{shalit2017}
Shalit, U., Johansson, F.D., Sontag, D.: Estimating individual treatment effect: generalization bounds and algorithms. In: 34th ICML. vol.~70. PMLR (2017)

\bibitem{shoush2024}
Shoush, M., Dumas, M.: Prescriptive process monitoring under resource constraints: A reinforcement learning approach. K{\"u}nstliche Intelligenz  \textbf{39} (2024)

\bibitem{shoush2021}
Shoush, M., Dumas, M.: Prescriptive process monitoring under resource constraints: A causal inference approach. In: Process Mining Workshops. vol.~433. Springer (2022)

\bibitem{shoush2022}
Shoush, M., Dumas, M.: When to intervene? prescriptive process monitoring under uncertainty and resource constraints. In: BPM Forum. vol.~458. Springer (2022)

\bibitem{shoushwhitebox}
Shoush, M., Dumas, M.: White box specification of intervention policies for prescriptive process monitoring. Data \& Knowledge Engineering  \textbf{155} (2025)

\bibitem{manela2024frugalflows}
de~Vassimon~Manela, D., Battaglia, L., Evans, R.J.: Marginal causal flows for validation and inference. In: 38th NeurIPS. vol.~37 (2024)

\bibitem{ppmremainingtime2019}
Verenich, I., Dumas, M., Rosa, M.L., Maggi, F.M., Teinemaa, I.: Survey and cross-benchmark comparison of remaining time prediction methods in business process monitoring. ACM Trans. Intell. Syst. Technol.  \textbf{10} (2019)

\bibitem{weytjens2023}
Weytjens, H., Verbeke, W., De~Weerdt, J.: Timed process interventions: Causal inference vs. reinforcement learning. In: BPM Workshops. vol.~492. Springer (2024)

\end{thebibliography}

\end{document}